*Discussion paper*

# Methods, Data, and Conceptual Change: Reflections from Two Quantitative Diachronic Case Studies

Catherine Wong[a]*, Bách Phan-Tất[b] and Susan Fitzmaurice[c]

[a]*School of History, Philosophy and Digital Humanities, University of Sheffield, Sheffield, United Kingdom;* [b]*Department of Linguistics, KU Leuven, Leuven, Belgium;* [c]*School of English, University of Sheffield, Sheffield, United Kingdom*

*Corresponding author: Catherine, Wong; catherine.wong@sheffield.ac.uk

## Author roles

For determining author roles, please use following taxonomy: https://credit.niso.org/

Please list the roles for each author

Catherine Wong: Conceptualization, Methodology, Investigation, Writing – original draft, Writing – review & editing.

Bách Phan Tất: Methodology, Investigation, Writing – original draft (Section 3), Writing – review & editing.

Susan Fitzmaurice: Conceptualization, Supervision, Writing – review & editing.

**Abstract**

This discussion paper reflects on how quantitative approaches to historical linguistics interact with dataset properties. Drawing on two worked examples, we examine English data using quad-based concept modelling of Early Modern English discourse in EEBO-TCP (c.1470s–1690s; ~765M words) alongside SynFlow analysis of scientific writing in Royal Society Corpus 6.0.4 (1750–1799; drawn from a ~78.6M-token open corpus). Through parallel comparison, the paper explores how each approach operationalises concepts, the data assumptions they entail, and the diachronic interpretations they support. We argue that comparative methodological reflection clarifies the limits of purely lexical, frequency-based approaches and highlights how dataset structure shapes the kinds of semantic change that quantitative methods can reliably detect.

Keywords: diachronic linguistics; historical English corpora; conceptual change; methodological comparison; quantitative methods

## (1) Introduction

Quantitative approaches have become increasingly prominent in diachronic linguistics, as large historical corpora and computational tools enable the analysis of linguistic patterns across extended time spans. However, the growing availability of historical datasets also raises an important methodological question: how do the structural properties of datasets shape the kinds of linguistic change that can be observed through quantitative analysis?

This discussion paper argues that diachronic change is method- and data- dependent. Rather than treating change as an inherent property of language itself that can be directly recovered from data, we suggest that what counts as observable change is also shaped by how datasets encode linguistic context and what analytical methods make visible. Different datasets encode linguistic structure in different ways, and computational approaches operationalise this structure through specific assumptions and preprocessing steps. As a result, methods do not simply analyse historical data, but actively shape the forms of change that become visible.

To examine this interaction, the paper presents two worked examples based on openly available historical English datasets. The first applies quad-based concept modelling applied to the Early English Books Online Text Creation Partnership corpus (EEBO-TCP), a large and heterogeneous collection of Early Modern English texts (c.1470s-1690s). The second uses SynFlow dependency-based analysis to examine syntactic co-occurrence patterns in a subset of the Royal Society Corpus (RSC) 6.0.4, focusing on eighteenth-century scientific writing (1750-1799). Each approach relies on different assumptions about linguistic structure and is applied to datasets with distinct properties, allowing us to explore how method-data interaction shapes analytical outcomes.

The aim of the paper is not to propose a new method or to provide a full empirical evaluation, but to offer a methodological reflection grounded in comparative analysis. By placing these approaches in parallel, we show how different methods foreground different dimensions of conceptual change and how dataset constraints condition what can be reliably modelled. In doing so, the paper contributes to ongoing discussions on transparency, reproducibility, and methodological fit in quantitative diachronic research.

More broadly, the paper highlights that quantitative analysis in historical linguistics is not method-neutral. Analytical choices are conditioned by the affordances and limitations of the data and interpretations of change must therefore be understood in relation to the interaction between method and dataset. Recognising this relationship is essential for developing more robust and context-sensitive approaches to the study of language change over time.

## (2) Case study 1: Quad-based Concept Modelling

This section presents the first case study, which examines quad-based concept modelling applied to EEBO-TCP. The aim is not to provide a comprehensive methodological evaluation, but to illustrate how dataset properties interact with computational methods to shape the kinds of patterns that become observable in historical corpora. The discussion therefore considers the properties of the dataset, the analytical assumptions of the method, and the types of conceptual structures that become visible through this interaction.

### (2.1) *Dataset properties: variation and heterogeneity in EEBO-TCP*

EEBO-TCP is a large openly available dataset of Early Modern English texts containing approximately 765 million words spanning the late fifteenth to the late seventeenth centuries (c.1470s-1690s) (Text Creation Partnership, 2020). The corpus consists of transcribed and encoded texts derived from the Early English Books Online (EEBO) collections, including materials catalogued in the Pollard and Redgrave and Wing short-title catalogues. It encompasses a wide range of early printed genres, including religious writings, political tracts, pamphlets, sermons, and literary works. As a result, EEBO-TCP represents a highly heterogeneous discursive environment in which different registers and domains of early modern public life intersect.

Like many historical corpora derived from early print sources, EEBO-TCP exhibits substantial orthographic variation and uneven linguistic regularity. Spelling variation, editorial normalisation practices, and the absence of consistently recoverable syntactic structure can complicate analyses that depend on stable lexical forms or reliable dependency relations. At the same time, the scale and diversity of the dataset make it particularly well suited to approaches that identify patterns through aggregated lexical co-occurrence across large textual environments. These characteristics favour analytical approaches that identify conceptual structure through aggregated lexical co-occurrence rather than through fine-grained syntactic relations. In this context, methodological choice is not simply a matter of analytical preference, but is constrained by the extent to which different forms of linguistic structure can be reliably recovered from the dataset. The next section outlines one such approach developed in the Linguistic DNA project (Fitzmaurice et al., 2017; Fitzmaurice & Mehl, 2022).

### (2.2) *Analytical approach: quad-based concept modelling*

Quad-based concept modelling, developed in the Linguistic DNA project, provides a computational approach for identifying discursive conceptual structures in historical English corpora. Rather than focusing on pairwise collocation, the method identifies sets of four strongly co-occurring lemmas, known as *quads*, within a moving window of textual context (typically 100 words). These quads capture recurrent patterns of lexical association that reflect how concepts are embedded within discursive environments (see Wong, Fitzmaurice, and Lam, under review).

The LDNA pipeline operates on normalised and lemmatised historical text, allowing spelling variation in the EEBO corpus to be standardised before quad extraction. This preprocessing step enables recurring lexical associations to be identified despite the orthographic variability typical of early printed materials.

By working with normalised and lemmatised text, and by filtering for meaningful lexical items, the method is designed to capture conceptual relationships that may remain obscured when using approaches based solely on frequency or pairwise collocation. The combination of normalisation and lemmatisation is therefore not only a technical step, but a necessary condition for enabling meaningful patterns of lexical association to emerge from a corpus characterised by substantial orthographical variability.

Building on this approach, the present project aggregates quads around focal lemmas to form larger networks of association, referred to as *constellations*. These constellations represent the broader discursive contexts in which particular terms occur, allowing conceptual patterns to emerge from repeated co-occurrence structures across large corpora. In heterogeneous historical datasets such as EEBO-TCP, this aggregation-based approach provides a way of modelling conceptual structure at the level of discourse rather than relying on fine-grained syntactic relations.

Under these conditions, aggregation-based modelling provides a robust means of capturing conceptual structure, as recurrent patterns of association can stabilise across diverse textual environments despite local variation and noise.

### (2.3) *Pattern stabilisation in heterogeneous textual environments*

When applied to a large and heterogeneous corpus such as EEBO-TCP, quad extraction reveals recurring patterns of lexical association that stabilise across diverse textual contexts. Rather than reflecting isolated collocations, the resulting quad structures highlight clusters of lemmas that repeatedly co-occur within discursive environments. Aggregated around focal terms, these clusters form constellations that provide a view of the conceptual terrain in which particular lexical items operate.

A brief illustration can be drawn from the comparison of the near-synonymous keywords *liberty* and *freedom* in Early Modern English discourse. Although the two terms frequently co-occur and appear interchangeable at the lexical level, their quad constellations reveal distinct patterns of association across the corpus. In this way, concept modelling highlights how conceptual structure can stabilise at the level of discourse even within a highly variable textual environment.

### (2.4) *What kind of conceptual change becomes visible*

The patterns revealed through quad-based concept modelling suggest that conceptual organisation in historical discourse can be observed to emerge from recurrent lexical association rather than from the behaviour of individual lexical items alone. In the EEBO-TCP corpus, aggregation of quad structures across diverse textual environments allow discursive conceptual configurations to be identified as stabilising despite substantial variation in spelling, genre, and linguistic form.

The comparison of *liberty* and *freedom* illustrates this dynamic. Although the two terms frequently appear interchangeable at the level of lexical usage, their respective constellations reveal distinct discursive environments within the corpus. In this sense, conceptual divergence becomes visible not through changes in individual words, but through the patterned associations that organise discourse around them. Methods that capture these aggregated structures therefore provide a way of modelling conceptual change that complements more traditional frequency-based approaches to historical semantics.

## (3) Case study 2: SynFlow analysis

This section uses SynFlow on a subset of the Royal Society Corpus (Fischer et al., 2020) as a worked example of method–dataset interaction. The aim is not to provide a full semantic case study, but to show how a dependency-based quantitative method depends on particular corpus properties and, in turn, makes certain dimensions of diachronic change more visible than others.

**(3.1)** ***Dataset properties (RSC subset)***

The Royal Society Corpus 6.0.4 (Fischer et al., 2020) is a diachronic corpus of English scientific writing based mainly on the *Philosophical Transactions* and *Proceedings* of the Royal Society of London. It offers a relatively coherent scientific genre over a long time span, with the full release covering around 47000 texts and 295.9 million tokens (1665–1996), and the open release covering around 78.6 million tokens (1665–1920). Its main strengths are genre continuity, diachronic depth, and metadata richness. In our demonstrative case study, we only use a subset of the Open version, limiting it to the 1750 – 1799 period and 3 topics, namely Biochemistry, Chemistry 1, Chemistry 2 (the topics are obtained from Menzel et al., 2021) . Our corpus contains only 913 texts and 351849 tokens, with a vocabulary size of 73628. Because SynFlow requires dependency parsed corpora, we re-parsed the dataset using Stanza (Qi et al., 2020), the state-of-the-art parser for English.

Upon qualitative inspection, we noticed some limitations of this corpus, which are also typical of historical corpora: OCR noise (*e.g., head – ach* instead of *headache*), historical spelling variations (e.g., *inconveniency* for *inconvenience*), and residual preprocessing error. Even though these problems have been documented and claimed to be 'optimised' throughout the development of the dataset (Kermes et al., 2016; Knappen et al., 2017), they are still present in our version, which is the latest release. These have negatively affected the parser, which is reflected in some wrong lemmatisations, wrong or unrecognised POS tags (labelled as X by Stanza), and wrong dependency relations. However, instead of trying to fix the problem, we want to see how using the original corpora and off-the-shell tagger affects our analysis.

**(3.2)** ***SynFlow***

SynFlow (Phan-Tất, 2025) builds on the principle of syntactic co-occurrence, according to which a concept is understood through the grammatical relations that structure its contextual behaviour.

In this framework, a concept is represented not by linear word proximity alone, but by the distribution of lexical fillers in dependency-based slots, such as adjectival modification, subjecthood, objecthood, and prepositional complementation. This slot-based representation makes it possible to trace how the internal semantic profile of a concept is distributed across different functional dimensions of use. When applied diachronically, SynFlow enables the comparison of these structured profiles across time, revealing not only whether a concept has changed, but also which semantic dimensions have shifted and how. In this sense, SynFlow aims to provide a more explainable and linguistically interpretable account of semantic change than approaches based solely on surface co-occurrence. Because it requires dependency parsing, it can only be applied to corpora which have already been parsed.

### (3.3) ***SynFlow x RSC***

Using SynFlow, we analysed the conceptual development of *air*, *water*, and *acid* in the RSC during the Chemical Revolution. Our findings indicate that SynFlow is able to capture the dynamic semantic profiles of all three concepts, uncovering historically interpretable patterns of conceptual stability, shift, and reorganisation. For example, the concept of *air* shifted from a general, undifferentiated substance to a chemically differentiated set of specific *airs*. More importantly, the effect of the known data issues appears to be minimal. In the high-frequency items and main dependency relations that are most relevant to the analysis, noise is limited, while errors associated with the target concepts tend to be sparse and readily identifiable. This suggests that, for this type of analysis, the original corpus data together with an off-the-shelf tagger can still yield reliable results.

However, caution is still required. Stanza is primarily trained on Modern/ Contemporary English, and eighteenth-century English remains relatively close to it. This likely helps explain why the tagging quality is still sufficient in our case. The same cannot be assumed for earlier historical periods, where the linguistic distance from Modern English is greater and the approach may no longer be reliable.

## (4) Comparative methodological discussion

The two case studies illustrate distinct but complementary interplay of dataset properties, analytical assumptions, and observable patterns of conceptual change. Rather than comparing methods in terms of performance, the discussion focuses on how each approach operationalises meaning within the constraints of its underlying dataset.

**(4.1)** ***Conceptualisation of meaning***

The two case studies operationalise the notion of 'concept' in fundamentally different ways, reflecting both the analytical assumptions of each method and the structural properties of the datasets to which they are applied. In the quad-based approach, conceptual structure is modelled through recurring patterns of lexical co-occurrence within a broad discursive window. Concepts are not treated as properties of individual lexical items, but as configurations of associations that emerge from repeated co-occurrence across heterogeneous textual environments. Meaning is therefore modelled as distributed across networks of lexical relations and stabilises at the level of discourse rather than at the level of individual tokens.

By contrast the SynFlow approach models conceptual structure through syntactic co-occurrence patterns derived from dependency relations. Here, meaning is operationalised in terms of the functional roles that lexical items occupy within structured linguistic environments. Concepts are inferred from shifts in these relational patterns over time, with syntactic dependencies providing a more fine-grained representation of how lexical items interact within specific domains of usage.

These differing operationalisations reflect not only methodological choices but also the constraints and affordances of the underlying datasets. In EEBO-TCP, where orthographic and stylistic variation, alongside  uneven linguistic regularity in grammatical and syntactic structures limit the reliability of syntactic analysis, concept modelling through aggregated lexical associations provides a robust way of capturing discursive structure. In the Royal Society Corpus, by contrast, the relative regularity of scientific prose enables dependency-based modelling of relational patterns, allowing conceptual change to be traced through shifts in syntactic co-occurence.

### (4.2) *Data assumptions and constraints*

The two approaches are grounded in different assumptions about the structure and reliability of the underlying data, which in turn shape the kinds of analytical procedures that can be applied. In the case of EEBO-TCP, the corpus is characterised by substantial orthographical variation, heterogeneous genre composition, and uneven grammatical and syntactic regularity across time. These properties limit the feasibility of approaches that depend on stable syntactic annotation or consistent surface forms. As a result, the LDNA pipeline relies on normalisation and lemmatisation to reduce variation at the lexical level, and on aggregation across large textual windows to identify recurring patterns of co-occurrence.

By contrast, the Royal Society Corpus provides a more controlled textual environment, consisting primarily of scientific prose with relatively consistent orthography and more regular syntactic structure. This makes it possible to apply dependency-based methods such as SynFlow, which rely on the identification of syntactic relations between lexical items. At the same time, the approach assumes the availability of sufficiently reliable part-of-speech tagging and dependency parsing, even when applied to historical data. This also means that for earlier historical stages or low-resource languages without reliable parsers, the method may be less applicable unless additional preprocessing, parser adaptation, or manual annotation is introduced.

These differences highlight how methodological choices are conditioned by the affordances and limitations of the datasets. In EEBO-TCP, where linguistic variability complicates fine-grained structural analysis, conceptual patterns are more robustly captured through aggregated lexical associations. In the Royal Society Corpus, the relative stability of syntactic structure allows conceptual relationships to be modelled through dependency-based representations, even though this introduces its own assumptions about the reliability of automated annotation.

### (4.3) *Observable dischronic patterns*

The interaction between dataset properties and analytical approach leads to different forms of diachronic pattern becoming visible in each case study. In EEBO-TCP, conceptual structure becomes visible through the stabilisation of recurring patterns of lexical association across heterogeneous textual environments. Despite substantial variation in spelling, genre, and linguistic form, aggregated co-occurrence patterns allow discursive conceptual configurations to persist and become identifiable at scale. Diachronic change, in this context, is observable in shifts in the relative prominence and composition of these constellations rather than in discrete changes to individual lexical items. As reported in Section 2, the distinction between *liberty* and *freedom* is reflected in different constellations of lexical associations, with *liberty* clustering around institutional and legal discourse and *freedom* around moral and spiritual contexts.

In the Royal Society Corpus, by contrast, diachronic patterns are captured through changes in syntactic co-occurrence within a more constrained domain of discourse. Dependency-based analysis makes it possible to trace how lexical items participate in evolving relational structures over time, revealing shifts in functional usage associated with developments in scientific discourse. As discussed in Section 3, for example, the reconceptualisation of the concept of *air* is realised in the several slots, most prominently the adjective slot. It shifts from being described in relatively general terms to being modified by more chemically specific adjectives. Here, conceptual change is observable through the reconfiguration of structured (syntactic) co-occurrence rather than through the aggregation of broad (window-based) co-occurrence patterns.

These case studies demonstrate that different methodological approaches foreground different dimensions of diachronic change. In one case, conceptual organisation becomes visible through the persistence and restructuring of discursive associations; in the other, it is revealed through shifts in relational patterns within more tightly structured linguistic environments. The contrast highlights how the form of observable change is shaped not only by the analytical method, but by the nature of the dataset to which it is applied.

### (5) Implications for quantitative diachronic research

The comparison of the two case studies highlights several implications for the use of quantitative methods in diachronic linguistics. First, the findings underscore the limitations of approaches that rely solely on lexical frequency or surface-level collaboration. While such methods can capture broad distributional trends, they may obscure underlying conceptual organisation that becomes visible through more complex patterns of association or relational structure. The case studies demonstrate that different methodological approaches can reveal distinct dimensions of semantic change, depending on how linguistic context is operationalised.

Second, the analysis emphasises the importance of aligning analytical methods with the structural properties of the dataset. As shown in the contrast between EEBO-TCP and the Royal Society Corpus, the reliability of particular forms of linguistic representation, whether lexical co-occurrence or syntactic dependency, depends on factors such as orthographic consistency, genre composition, and the feasibility of preprocessing. Methodological choices are therefore not neutral, but conditioned by the affordances and limitations of the data.

Third, the discussion highlights the need for greater methodological transparency in quantitative diachronic research. Making explicit the assumptions underlying data preprocessing, annotation, and analytical modelling is essential for the interpretation and reproducibility of results, particularly when working with historical corpora characterised by variation and uncertainty. This is especially important in the context of open datasets, where reuse depends not only on data availability but also on clear documentation of how data has been transformed and analysed.

Finally, the paper points to the value of comparative, example-based approaches for methodological reflection. Rather than seeking a single optimal method, placing different analytical strategies in relation to one another makes it possible to identify how each approach foregrounds particular aspects of linguistic change. This perspective supports a more flexible and context-sensitive understanding of quantitative analysis, in which methods are evaluated in relation to the datasets they are designed to interrogate.

## (6) Conclusion

This paper argues that diachronic change is method- and data-dependent, and demonstrates this through two contrasting computational approaches. Diachronic change is not directly observable as an inherent property of language alone, but is shaped by the interaction between dataset properties and analytical methods. Through two case studies, we have shown that different approaches make different aspects of conceptual change visible, depending on how linguistic structure is operationalised and what forms of data are available.

In the case of EEBO-TCP, the heterogeneity and orthographical variability of early modern texts favour aggregation-based approaches that capture stabilised patterns of lexical association at the level of discourse. In contrast, the relative syntactic regularity of the Royal Society Corpus enables dependency-based modelling of relational patterns, allowing conceptual change to be traced through shifts in structured co-occurrence. These differences do not simply reflect methodological preference, but arise from the constraints and affordance of the underlying datasets.

The comparison highlights that quantitative approaches to historical linguistics do not merely analyse pre-existing patterns of change, but actively shape what counts as observable change. Methodological transparency, including explicit attention to data preprocessing, annotation, and analytical assumptions, is therefore essential for the interpretation and resume of historical datasets. By foregrounding the relationship between method and data, this paper contributes to a more context-sensitive understanding of quantitative diachronic analysis, in which different approaches are evaluated in relation to the datasets they are designed to interrogate.

**Acknowledgements**

This research builds on the Linguistic DNA project, Modelling concepts and semantic change in English 1500–1800, funded by the Arts and Humanities Research Council (AHRC AH/M00614X/1). The project was led by Susan Fitzmaurice (Principal Investigator), with Co-Investigators Michael Pidd, Justyna Robinson, and Marc Alexander, and was supported by research associates Fraser Dallachy, Iona Hine, and Seth Mehl. Technical development was carried out by Matthew Groves, Katherine Rogers, and Brian Aitken. Further information is available at http://linguisticdna.org (accessed 1 April 2026).

**Funding statement**

This research draws on work supported by the Arts and Humanities Research Council (AHRC AH/M00614X/1).

**Competing interests**

The authors have no competing interests to declare.